\documentclass[sigconf]{acmart}

\usepackage[font=rm]{caption}

\AtBeginDocument{%
  }

\setcopyright{acmlicensed}
\copyrightyear{2018}
\acmYear{2018}
\acmDOI{XXXXXXX.XXXXXXX}
\acmConference[Conference acronym 'XX]{Make sure to enter the correct
  conference title from your rights confirmation email}{June 03--05,
  2018}{Woodstock, NY}
\acmISBN{978-1-4503-XXXX-X/2018/06}

\newcommand{\Patk}[1]{\mbox{\Pat@$#1$}}
\newcommand{\RBPatp}[1]{\mbox{\RBP@$#1$}}

\newcommand{\NDCGatk}[1]{\mbox{\NDCG@$#1$}}
\newcommand{\ERRatk}[1]{\mbox{\ERR@$#1$}}

\newcommand{\myurl}[1]{{\url{#1}}}

\newcommand{\myparagraph}[1]{\vspace{0.2\baselineskip}\noindent{\textbf{#1}}.~}

\newcommand{\mycomment}[1]{}

\newlength{\onedigit}
\newcounter{todocount}
\begin{document}

%%
%% The "title" command has an optional parameter,
%% allowing the author to define a "short title" to be used in page headers.
\title{Fault2Flow: An AlphaEvolve-Optimized Human-in-the-Loop Multi-Agent System for Fault-to-Workflow Automation}

%%
%% The "author" command and its associated commands are used to define
%% the authors and their affiliations.
%% Of note is the shared affiliation of the first two authors, and the
%% "authornote" and "authornotemark" commands
%% used to denote shared contribution to the research.
\author{Yafang Wang}
\affiliation{%
  \institution{SGITG Accenture Information Technology}
    \state{Beijing}
  \country{China}
}

\author{Yangjie Tian}
\affiliation{%
  % \institution{Institute for Sustainable Industries and Liveable Cities, Victoria University}
  \institution{ISILC, Victoria University}
  \institution{Kexin Melbourne AI Research Center}
  \state{Melbourne}
  \country{Australia}
}

\author{Xiaoyu Shen}
\affiliation{%
  \institution{Eastern Institute of Technology}
  \state{Ningbo}
  \country{China}
}

\author{Gaoyang Zhang}
\affiliation{%
  \institution{SGITG Accenture Information Technology}
  \state{Beijing}
  \country{China}
}

% \author{Jiaze Sun,He Zhang}
% \affiliation{%
%   \institution{Kexin Research Institute}
%   \state{Beijing}
%   \country{China}
% }

\author{Jiaze Sun,He Zhang,Ruohua Xu}
\authornotemark[1]
\affiliation{%
  \institution{CNPIEC Kexin Technology}
  \state{Beijing}
  \country{China}
}

\author{Feng Zhao}
\authornotemark[1]
\affiliation{%
  \institution{SGITG Accenture Information Technology}
  \state{Beijing}
  \country{China}
}

% \author{
% 	\uppercase{Zining Liu}\authorrefmark{1},
% 	\uppercase{Chong Long\authorrefmark{2}, Xiaolu Lu\authorrefmark{3}, Zehong Hu\authorrefmark{4}, Jie Zhang\authorrefmark{5}, Yafang Wang\authorrefmark{2}}}

% \address[1]{School of Software, Shandong University, Jinan 250101, Shandong, China(e-mail: liuzining@mail.sdu.edu.cn)}
% \address[2]{Ant Financial Services Group, Z Space No. 556 Xixi Road Hangzhou,  310099, Zhejiang, China(e-mail: \{huangxuan.lc;yafang.wyf\}@antfin.com)}
% \address[3]{RMIT University, GPO Box 2476, Melbourne VIC 3001 Australia (e-mail: xiaolu.lu@rmit.edu.au )}
% \address[4]{Alibaba Group, 969 West Wen Yi Road, Hangzhou 311121, Zhejiang, China(e-mail: HUZE0004@e.ntu.edu.sg )}
% \address[5]{School of Computer Science and Engineering, Nanyang Technological University, 50 Nanyang Avenue, Singapore, 639798(e-mail: zhangj@ntu.edu.sg)}
%%
%% By default, the full list of authors will be used in the page
%% headers. Often, this list is too long, and will overlap
%% other information printed in the page headers. This command allows
%% the author to define a more concise list
%% of authors' names for this purpose.
\renewcommand{\shortauthors}{Yafang Wang et al.}

%%
%% The abstract is a short summary of the work to be presented in the
%% article.
\begin{abstract}
Power grid fault diagnosis is a critical process hindered by its reliance on manual, error-prone methods. Technicians must manually extract reasoning logic from dense regulations and attempt to combine it with tacit expert knowledge, which is inefficient, error-prone, and lacks maintainability as ragulations are updated and experience evolves. While Large Language Models (LLMs) have shown promise in parsing unstructured text, no existing framework integrates these two disparate knowledge sources into a single, verified, and executable workflow. To bridge this gap, we propose Fault2Flow, an LLM-based multi-agent system. Fault2Flow systematically: (1) extracts and structures regulatory logic into PASTA-formatted fault trees; (2) integrates expert knowledge via a human-in-the-loop interface for verification; (3) optimizes the reasoning logic using a novel AlphaEvolve module; and (4) synthesizes the final, verified logic into an n8n-executable workflow. Experimental validation on transformer fault diagnosis datasets confirms 100\% topological consistency and high semantic fidelity. Fault2Flow establishes a reproducible path from fault analysis to operational automation, substantially reducing expert workload. 
%The demo system recording could be found by the link~\url{https://note.kxsz.net/share/c4852760-b164-4179-9c6e-b1aa05c0e79f}.
\end{abstract}

%%
%% The code below is generated by the tool at http://dl.acm.org/ccs.cfm.
%% Please copy and paste the code instead of the example below.
%%

\begin{CCSXML}
<ccs2012>
  <concept>
    <concept_id>10010147.10010178.10010179</concept_id>
    <concept_desc>Computing methodologies~Multi-agent systems</concept_desc>
    <concept_significance>500</concept_significance>
  </concept>
  <concept>
    <concept_id>10010147.10010257.10010258.10010259</concept_id>
    <concept_desc>Computing methodologies~Reasoning about beliefs and knowledge</concept_desc>
    <concept_significance>300</concept_significance>
  </concept>
  <concept>
    <concept_id>10010147.10010257.10010258.10010261</concept_id>
    <concept_desc>Computing methodologies~Reinforcement learning</concept_desc>
    <concept_significance>300</concept_significance>
  </concept>
  <concept>
    <concept_id>10003120.10003121.10003122.10003334</concept_id>
    <concept_desc>Human-centered computing~User studies</concept_desc>
    <concept_significance>300</concept_significance>
  </concept>
  % <concept>
  %   <concept_id>10003033.10003083.10003095</concept_id>
  %   <concept_desc>Networks~Network reliability</concept_desc>
  %   <concept_significance>300</concept_significance>
  % </concept>
</ccs2012>
\end{CCSXML}

\ccsdesc[500]{Computing methodologies~Multi-agent systems}
\ccsdesc[300]{Computing methodologies~Reasoning about beliefs and knowledge}
% \ccsdesc[300]{Computing methodologies~Reinforcement learning}
\ccsdesc[300]{Human-centered computing~User studies}
% \ccsdesc[300]{Networks~Network reliability}

%%
%% Keywords. The author(s) should pick words that accurately describe
%% the work being presented. Separate the keywords with commas.
\keywords{AlphaEvolve Optimization; Human-in-the-loop; Multi-agent system; Fault Tree Analysis (FTA); Workflow automation}
%% A "teaser" image appears between the author and affiliation
%% information and the body of the document, and typically spans the
%% page.

% \begin{teaserfigure}
%   \includegraphics[width=\textwidth]{sampleteaser}
%   \caption{Seattle Mariners at Spring Training, 2010.}
%   \Description{Enjoying the baseball game from the third-base
%   seats. Ichiro Suzuki preparing to bat.}
%   \label{fig:teaser}
% \end{teaserfigure}

\received{20 February 2007}
\received[revised]{12 March 2009}
\received[accepted]{5 June 2009}

%%
%% This command processes the author and affiliation and title
%% information and builds the first part of the formatted document.
\maketitle

\section{Introduction}

Power grid equipment fault diagnosis is a critical process that ensures the reliability and safety of power systems. When abnormal conditions arise, such as transformer overheating, insulation failure or partial discharge, technicians must determine the root cause, assess possible propagation paths, and choose appropriate mitigation actions. This diagnosis is typically guided by two disconnected knowledge sources:
(1) Regulatory documents, which specify standardized fault scenarios, safety requirements, and mandatory handling procedures, and
(2) Technicians’ experiential knowledge, accumulated over years of on-site troubleshooting. Traditionally, fault diagnosis relies heavily on manual interpretation of these documents and expert experience. Engineers read through lengthy regulations, extract conditional logic by hand, and encode fault-handling rules into tools or checklists~\cite{li2010method, han2017power}. This manual approach suffers from three major drawbacks: it leads to incomplete knowledge (as expert experience is hard to codify), poor maintainability, and severe inefficiencies. For instance, any regulatory revision requires engineers to manually find and update the corresponding logic, a cumbersome process that can introduce new errors. Addressing these challenges demands resolving two core issues: first, the \emph{automatic extraction of fault reasoning logic from unstructured regulations}, and second, the creation of \emph{user-friendly tools for the efficient accumulation of experiential knowledge}. Furthermore, these two knowledge types must be structurally integrated and verified to enhance both extraction accuracy and tool practicality.

Recent technological advancements lay a solid foundation for tackling these problems. Large language models (LLMs) have enabled a new ``knowledge + data dual-driven'' paradigm, moving beyond structured data. This allows them to parse semantic logic directly from unstructured text, as demonstrated in fields from healthcare~\cite{zhu2024text2mdt,li2025ftmdt} to law~\cite{han2024legalasst}. Beyond simple extraction, related work provides a basis for high-reliability reasoning. Generative AI has been used to automate formal Fault Tree Analysis (FTA)~\cite{sudhir2024fta}, and methods like Logic Augmented Generation (LAG)~\cite{gangemi2025logic} use knowledge graphs to ensure LLMs operate within strict logical boundaries. Within the power grid domain itself, multi-agent architectures like Grid-Agent~\cite{zhang2025grid} and GridMind~\cite{jin2025gridmind} have proven effective for enhancing modularity and interpretability by decomposing complex tasks.

However, no existing system integrates these two knowledge types (formal regulations and expert insights) or verifies them within a single, automated workflow. To bridge this gap, we propose \textbf{Fault2Flow}, an LLM-based multi-agent system that addresses this end-to-end challenge. Fault2Flow systematically (1) \emph{Extracts and Structures Regulatory Logic}, automatically parsing regulatory documents and creating PASTA-formatted fault trees; (2) \emph{Integrates Expert Knowledge} through a human-in-the-loop interface for expert review, refinement, and verification; (3) \emph{Optimizes Reasoning} using a novel AlphaEvolve module that refines fault tree structures; and (4) \emph{Automates Execution}, synthesizing the verified logic into an n8n-executable workflow. Experimental validation on transformer fault diagnosis datasets confirms 100\% topological consistency, full reachability coverage, and high semantic fidelity. By combining modular agent cooperation with AlphaEvolve optimization, Fault2Flow substantially reduces expert workload and establishes a reproducible path from fault analysis to operational automation.

\section{Fault2Flow}

\begin{figure}[h]
  \centering
  \includegraphics[width=\linewidth]{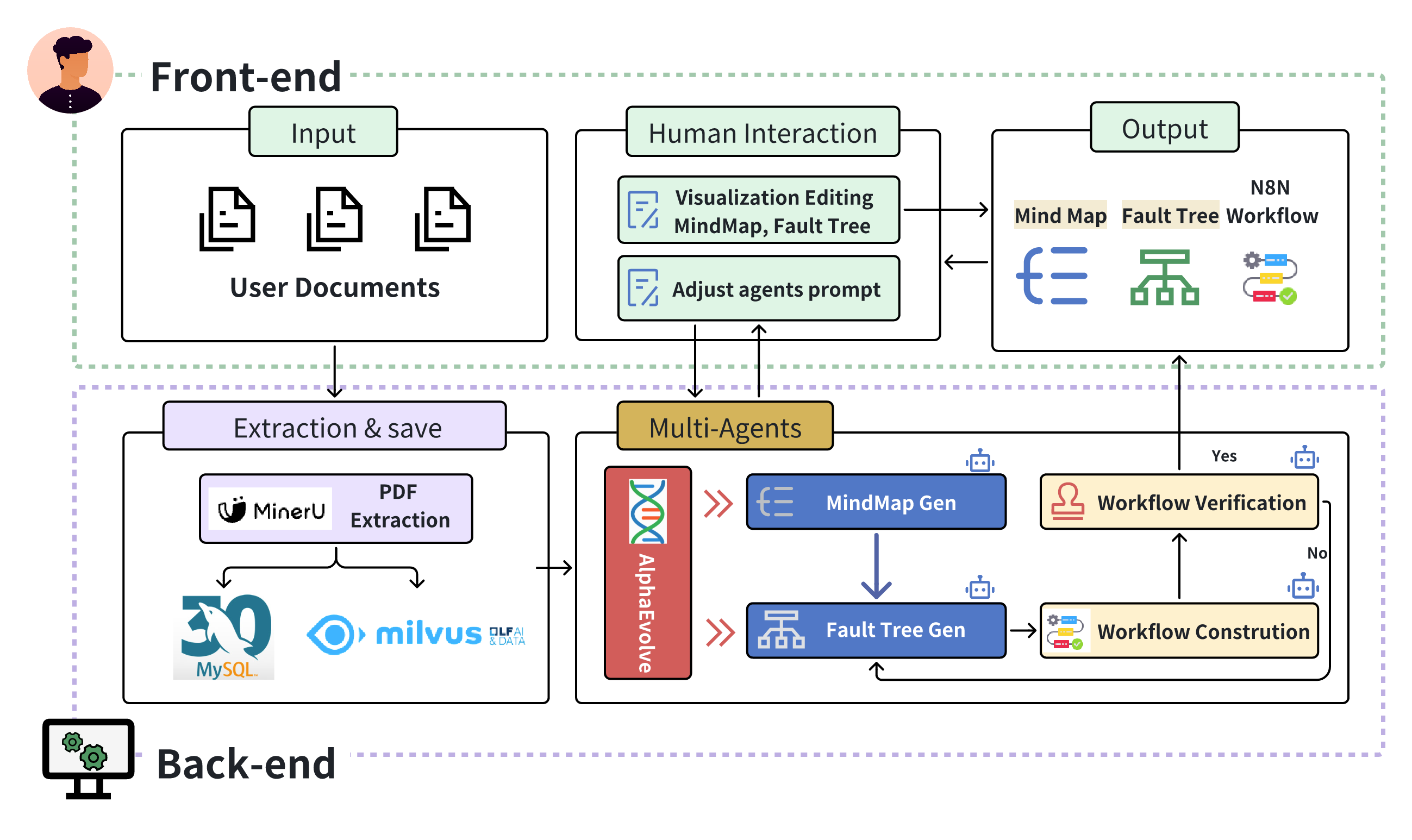}
  \caption{\small Overall architecture of the Fault2Flow system}
  \label{fig: Fault2Flow design}
\end{figure}

The overall architecture of Fault2Flow is illustrated in Figure~\ref{fig: Fault2Flow design}. It follows a decoupled front-end/back-end design: the front end provides interactive tools for domain experts, while the back end implements a multi-agent reasoning pipeline that automates fault-logic extraction and workflow generation. At a high level, Fault2Flow operates through three coordinated stages. (1) Document Understanding: regulatory PDFs are parsed and converted into structured Markdown. (2) Knowledge Structuring: a sequence of LLM agents transforms the Markdown into (i) mind maps, which organize concepts, conditions, and decision rules into a hierarchical structure, and (ii) fault trees, which formalize causal and logical relationships in a machine-readable format. (3) Workflow Synthesis and Verification: the structured representations are translated into n8n workflow definitions, which are then automatically validated using synthesized test cases to ensure correctness and consistency.

\begin{figure}[h]
  \centering
  \includegraphics[width=\linewidth]{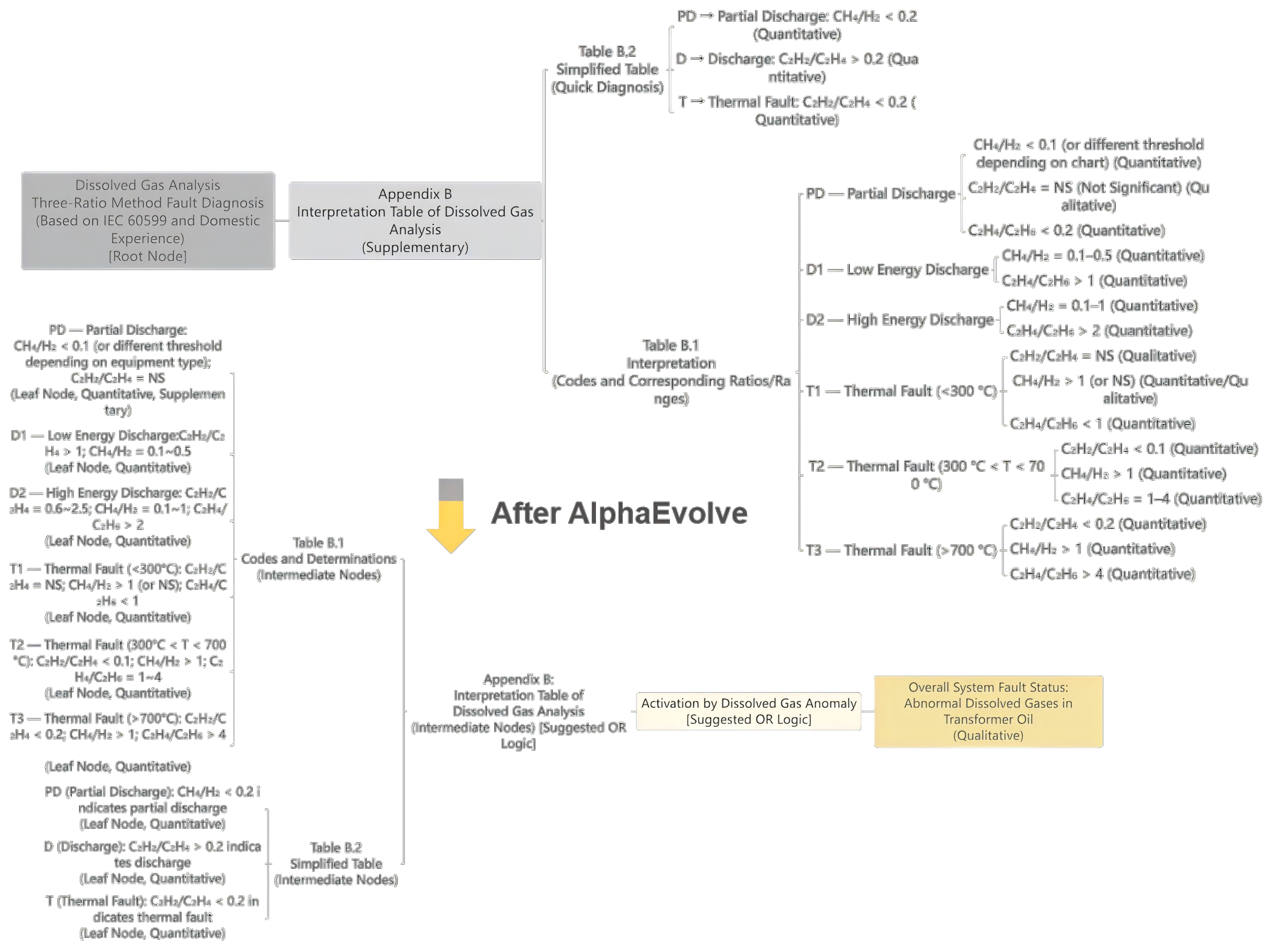}
  \caption{\small Mind-map before and after AlphaEvolve optimization}
  \label{fig: mind map compar}
\end{figure}

\subsection{Frontend Design}
Built on Vue3 and Element Plus, the front end provides a user-friendly interface for document upload, mind map/fault tree editing, and workflow initiation. It integrates pdfjs-dist\footnote{https://www.npmjs.com/package/pdfjs-dist} (PDF preview), xlsx\footnote{https://www.npmjs.com/package/xlsx} (table parsing), and ECharts\footnote{https://echarts.apache.org/zh/index.html} (visualization) to support practical engineering use.

A key design principle is \textit{human-in-the-loop} oversight. As shown in Figure~\ref{fig: Fault2Flow design}, domain experts can review intermediate outputs—mind maps, fault trees, and workflow configurations—providing corrections or refinement instructions when necessary. This mechanism ensures that the multi-agent pipeline remains aligned with industry standards and domain-specific requirements.

\subsection{Backend Design}

The backend, implemented in Python 3, handles PDF parsing, multi-agent reasoning, structural optimization, and workflow generation. It uses MinerU\footnote{https://mineru.net/} for high-quality PDF extraction, converting user-uploaded regulatory documents into structured Markdown. The processed text is stored in MySQL for reliable structured data management and indexing, while vector embeddings are stored in Milvus\footnote{https://milvus.io/} to support semantic vector search and similarity-based retrieval. This dual-storage design enables efficient document-level access and semantic-level retrieval throughout the pipeline.

The multi-agent workflow consists of the following components:

\myparagraph{PDF Extraction}
MinerU is an open-source extractor designed to process multi-modal PDFs containing text, tables, formulas, and images. It converts regulatory documents into Markdown, providing clean, structured input for downstream reasoning tasks.

\myparagraph{Mind Map Generation Agent}
Taking the parsed Markdown as input, the LLM is prompted via a one-shot example to produce a PlantUML-formatted mind map that captures both qualitative descriptions and quantitative decision criteria in the regulation~\cite{brown2020language}. Experts may then refine the mind map either by editing it directly or by supplying additional instructions to the LLM.

\myparagraph{Fault Tree Translation Agent}
Once the mind map is validated, it is used to guide an LLM-based translation step that converts its quantitative logic into PASTA-formatted fault tree code. As with the previous stage, experts can revise the output manually or through corrective prompts. The leaf nodes of the PASTA fault tree form the basis for determining workflow input parameters in later stages.

\myparagraph{AlphaEvolve Optimization}
To enhance structural coherence and logical consistency in both mind maps and fault trees, \textbf{Fault2Flow} incorporates the AlphaEvolve evolutionary optimizer\footnote{https://github.com/algorithmicsuperintelligence/openevolve}
. AlphaEvolve employs a multi-island evolutionary strategy where multiple subpopulations evolve independently while periodically exchanging elite individuals. In each iteration, a parent program and several “inspiration” programs are sampled from the elite archive and composed into a joint prompt, prompting the LLM to generate improved code. New candidates are evaluated for readability and logical validity; the strongest solutions replace weaker individuals, gradually improving the structure over successive iterations (Figure~\ref{fig: mind map compare}).

\myparagraph{Workflow Construction Agent}
The construction module combines LLM-based extraction with deterministic rule-based logic. First, the PASTA code is parsed into a tree structure supporting depth-first traversal. Next, each leaf node is processed by prompting the LLM to extract workflow input parameters according to n8n trigger specifications. Using a post-order traversal strategy, the system then generates and links n8n node configurations, and finally invokes the n8n API to activate the executable workflow.

\myparagraph{Workflow Verification Agent}
To ensure correctness, the system prompts the LLM to synthesize test samples and expected results (indicating whether a fault should be triggered). These test cases are automatically executed against the generated n8n workflow. If outputs match expectations, verification succeeds. Otherwise, the failure traces are fed back to the model to guide refinement and regenerate the workflow. This iterative loop continues until successful validation or reaching the maximum number of iterations.

\begin{figure}[h]
  \centering
  \includegraphics[width=\linewidth]{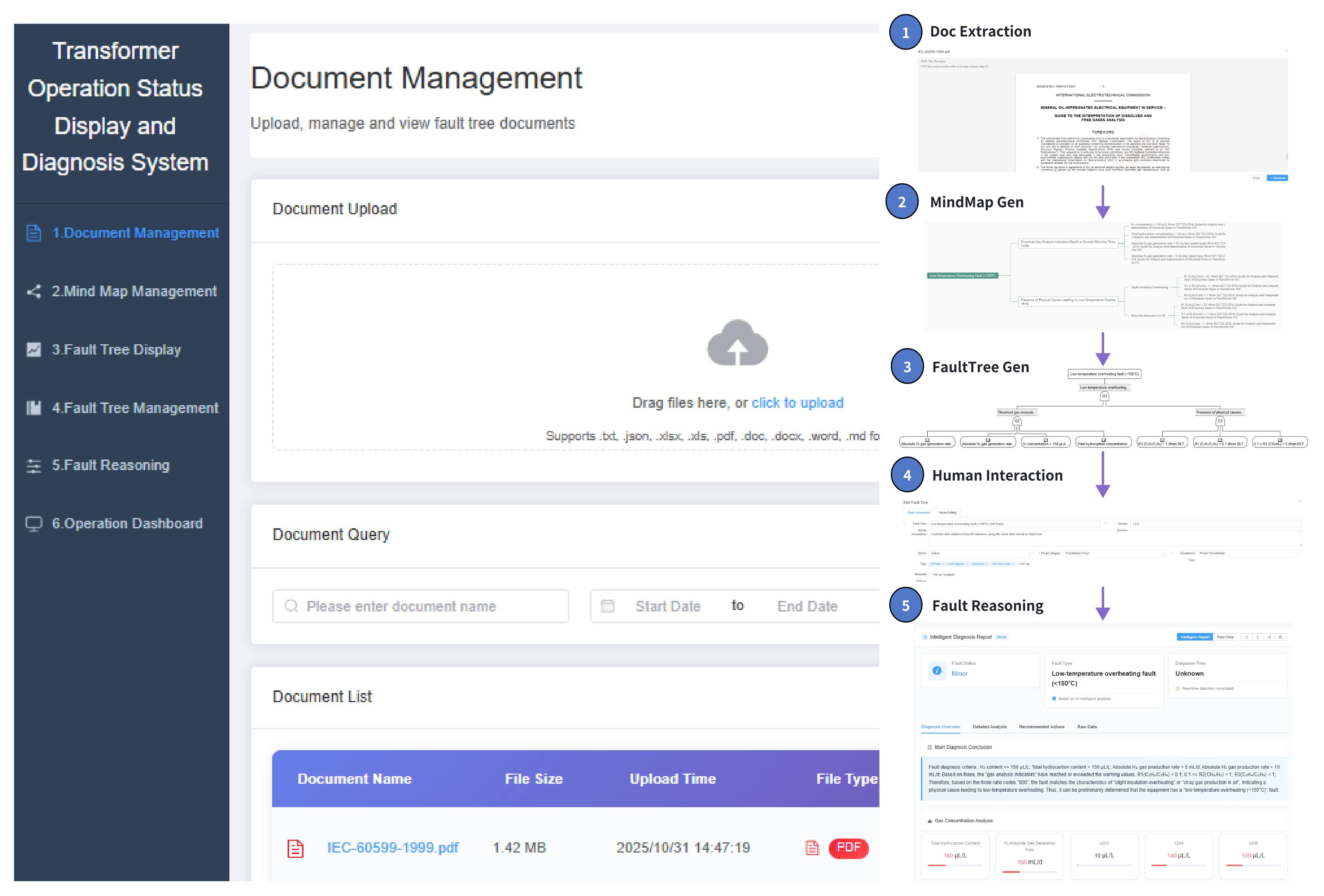}
  \caption{\small Case study demonstrating the transformation from reports to executable n8n workflow in three-ratio diagnostic.}
  \label{fig:case-study}
\end{figure}
\section{Demonstration}

To demonstrate the end-to-end capabilities of Fault2Flow, we conducted a case study on transformer fault diagnosis. The source material for this demonstration was the national standard ``DL/T 722-2014: Guidelines for Dissolved Gas Analysis and Judgement of Transformer Faults.'' As shown in Figure~\ref{fig:case-study}, this case study showcases the full pipeline, from parsing the raw document to generating a verified, executable workflow.

\myparagraph{Data Upload and Parsing} The experiment focused on the ``three-ratio method'' described in the standard for dissolved gas analysis (DGA). The uploaded document was first processed by MinerU, which automatically extracted the hierarchical text, diagnostic tables, and numerical thresholds. This step produced a structured Markdown corpus, making the complex regulatory logic accessible for the subsequent reasoning agents.

\myparagraph{Mind Map and Fault-Tree Generation}The The \textit{Mind-Map Generation Agent} then used this corpus to generate a PlantUML-formatted mind map. This agent, using the \textit{Qwen3-Next-80B-A3B-Instruct} model (at \texttt{temperature = 0.0}) in a one-shot configuration, produced an initial structure that was further refined by our \textit{AlphaEvolve} module. The resulting diagram accurately captured the full diagnostic hierarchy of the three-ratio method, including both quantitative ratios (C$_2$H$_2$/C$_2$H$_4$, CH$_4$/H$_2$, C$_2$H$_4$/C$_2$H$_6$) and the corresponding qualitative fault patterns (e.g., overheating, partial discharge, arcing discharge).

After an expert reviewed and approved this mind map, the \textit{Fault-Tree Translation Agent} converted the PlantUML structure into \textit{PASTA DSL}. This step produced a machine-readable logical tree where each quantitative threshold was modeled as a component (e.g., \texttt{C$_2$H$_2$/C$_2$H$_4$ < 0.1}) and intermediate nodes represented diagnostic gates (e.g., \textit{Low-temperature overheating < 150 °C}). The output successfully passed all PASTA self-checking procedures, confirming its syntactic correctness and logical integrity. 

\myparagraph{Workflow Synthesis and Verification} Finally, the \textit{Workflow Construction Agent} traversed the verified PASTA code to automatically synthesize an \textit{n8n executable workflow}. It mapped each leaf node to a corresponding \texttt{n8n} input form and transformed the logical gates into conditional \texttt{IF/ELSE} branches.
Subsequently, the \textit{Workflow Verification Agent} took over. It automatically generated synthetic test data and expected outputs for each fault type, then executed the workflow iteratively. This verification loop confirmed that all diagnostic paths were reachable and that the workflow's behavior was semantically identical to the logic in the original PASTA model.

Overall, this case study demonstrates the capability of \textbf{Fault2Flow} to bridge the gap from complex, domain-specific fault logic (in unstructured text) to automated, verifiable operational workflows in a real-world power grid scenario.

\section{Evaluation}

 To validate the effectiveness of the proposed framework, we conducted a comprehensive evaluation across multiple LLMs focusing on the \emph{workflow generation} stage, which forms the core of the fault-to-workflow transformation pipeline. All experiments were conducted under identical inputs to ensure fair comparison.

\myparagraph{Data} Our experiments used a transformer fault-diagnosis dataset with 16 cases (9 via three-ratio method, 7 via characteristic gas method). Each case includes an expert-designed fault tree (converted to PASTA code by engineers), forming 16 test samples (two diagnostic categories) with PASTA representations and auto-generated n8n workflows.

\myparagraph{Evaluation Metrics} We designed four complementary metrics to assess the system performance:  

\textbf{(1) Logical Readability and Maintainability (LRM)}:
This metric evaluates the interpretability and modular clarity of the generated workflows. GPT-5 was employed to emulate expert evaluation, simulating engineers familiar with both PASTA and n8n. The model scored the conversion results using a 5-point Likert scale (1 = disorganized and unreadable; 5 = highly readable and modular). The final normalized LRM score represents the aggregated assessment of logical clarity, hierarchical organization, and naming consistency.

\textbf{(2) Semantic Fidelity (SF)}:
Semantic fidelity measures whether the generated workflow fully preserves the logical semantics of the original PASTA representation, including nested conditions and exception handling. We employed GPT-5 to evaluate each workflow by comparing its execution output with the expected diagnostic reasoning, to identify semantic deviations. The score is normalized to the range $[0, 1]$, where $1$ indicates perfect semantic consistency.

\textbf{(3) Topological Consistency (TC)}:
Topological consistency measures the structural correspondence between the PASTA fault tree and the generated n8n workflow. It is computed as:
\begin{equation}
TC = \frac{|E_{PASTA} \cap E_{n8n}|}{|E_{PASTA}|},
\end{equation}
where $E_{PASTA}$ and $E_{n8n}$ denote the directed edge sets of the PASTA logic graph and n8n workflow graph. This metric reflects how faithfully control and data flows are preserved during translation.

\textbf{(4) End-to-End Reachability Coverage (E2ERC)}:
This metric evaluates whether all logical paths in the original PASTA model remain executable in the generated workflow. It is defined as:
\begin{equation}
E2ERC = \frac{|\mathcal{P}_{cov}|}{|\mathcal{P}_{ref}|},
\end{equation}
where $\mathcal{P}_{ref}$ is the set of valid paths in the original fault tree, and $\mathcal{P}_{cov}$ is the subset of reachable paths in the generated workflow. A higher E2ERC means better coverage and functional completeness.

\myparagraph{Results and Discussion} As shown in Table \ref{tab:evaluation_results}, the \textbf{Fault2Flow} framework significantly outperforms end-to-end baseline models that directly convert text into workflows, showing clear improvements in \textbf{TC}, \textbf{E2ERC}, and overall success rate. In contrast, baseline models exhibit incomplete topology mapping ($\mathrm{TC}<0.94$) and lower reachability coverage ($\mathrm{E2ERC}<0.72$), further confirming the necessity of the structured reasoning and verification mechanisms integrated into the \textbf{Fault2Flow}. Moreover, models based on the Fault2Flow framework consume only about one-tenth of the computational resources required by end-to-end approaches.
\begin{table}[!htp]
\centering
\caption{\small Evaluation results of different LLMs configurations on the transformer fault-diagnosis dataset.}
\label{tab:evaluation_results}
 \resizebox{0.5\textwidth}{!}{ %
\begin{tabular}{lcccccc}\toprule

 \textbf{Model} & \textbf{LRM} & \textbf{SF} & \textbf{TC} & \textbf{E2ERC}  & \textbf{Succ./Fail.} &\textbf{Avg. Tokens}\\\midrule
  Qwen3-80B(End2End)& 0.80& 0.89& 0.81& 0.69&14/2 &21334.21\\
 GPT-4.1(End2End)& 0.77& \textbf{0.92}& 0.93& 0.71&12/4 &19668.83\\
  \hline
  Qwen3-14B (Fault2Flow)& 0.79& 0.9& \textbf{1.0}& \textbf{1.0}&\textbf{16/0} &1868.56\\
  Qwen3-32B (Fault2Flow)& \textbf{0.81}& 0.9& \textbf{1.0}& \textbf{1.0}&\textbf{16/0} &1879.81\\
  Qwen3-80B(Fault2Flow)& 0.80& 0.9& \textbf{1.0}& \textbf{1.0}&\textbf{16/0} &\textbf{1799.83}\\

 GPT-4.1 (Fault2Flow) & 0.79 & 0.9 & \textbf{1.00} & \textbf{1.0}  &\textbf{16/0} &1887.62\\ \bottomrule
\end{tabular}
}
\end{table}

\section{Conclusion}
This paper presented \textbf{Fault2Flow}, an AlphaEvolve-optimized human-in-the-loop multi-agent system that bridges fault reasoning and workflow execution in power grids. By integrating symbolic Fault Tree Analysis with LLM-driven inference, the system achieves interpretable and verifiable automation from unstructured diagnostic reports to executable n8n workflows. Experimental results on transformer fault diagnosis demonstrate the reliability of the AlphaEvolve optimization and cooperative agents. Future work will extend the framework to broader grid scenarios and incorporate real-time feedback for adaptive workflow refinement.

\section{Ethical Use Of Data}
Our agent evaluation dataset was collected via human annotation, with participants informed about the research and data usage.

%%
%% The acknowledgments section is defined using the "acks" environment
%% (and NOT an unnumbered section). This ensures the proper
%% identification of the section in the article metadata, and the
%% consistent spelling of the heading.
\begin{acks}
We thank Jinyao Li for providing the manual transformer fault-diagnosis dataset and Yongsong Li for his research on AlphaEvolve, mind map, and fault tree generation. We also thank Guofeng Wang for the demo front-end development, and He Yang and Shanghong Zou for agent development and testing.
\end{acks}

%%
%% The next two lines define the bibliography style to be used, and
%% the bibliography file.
\bibliographystyle{ACM-Reference-Format}
\bibliography{references}

\end{document}